\documentclass{article}

\PassOptionsToPackage{numbers,sort&compress}{natbib}
\usepackage[preprint]{neurips_2026}

\usepackage[utf8]{inputenc}
\usepackage[T1]{fontenc}
\usepackage{hyperref}
\usepackage{url}
\usepackage{booktabs}
\usepackage{amsmath}
\usepackage{amssymb}
\usepackage{amsfonts}
\usepackage{nicefrac}
\usepackage{microtype}
\usepackage{xcolor}
\usepackage{graphicx}
\usepackage{enumitem}
\usepackage{multirow}
\usepackage{makecell}
\usepackage{algorithm}
\usepackage{algorithmic}
\usepackage{float}
\usepackage{xspace}

\newcommand{\ourmethod}{\textsc{Conf-KV}\xspace}
\newcommand{\ourmethodq}{\textsc{Conf-KV+INT8}\xspace}

\title{\ourmethod: Confidence-Aware KV Cache Eviction with Mixed-Precision Storage for Long-Horizon LLM Inference}

\author{
  Yubo Li \hspace{2em}
  Yidi Miao \hspace{2em}\\
  \texttt{\{yubol, yidim\}@andrew.cmu.edu} \\
  Carnegie Mellon University
}

\begin{document}

\maketitle
\begingroup
\renewcommand{\thefootnote}{}
\footnotetext{Code will be released upon acceptance.}
\addtocounter{footnote}{-1}
\endgroup

\begin{abstract}
Long-horizon LLM inference turns the key--value (KV) cache into the dominant GPU memory consumer and makes per-token attention increasingly expensive. Many common eviction policies use static recency windows or historical attention, leaving unused a signal computed on every decoding step: the model's current uncertainty. We introduce \ourmethod, a KV-cache manager that converts the next-token distribution into a scalar confidence score and uses it to choose the per-step cache budget, retaining more context when the model is uncertain and pruning aggressively when it is confident. Within each budget, tokens are ranked by a composite of accumulated attention mass and recency, while a protected recent window preserves local coherence. We combine the policy with blockwise online-softmax attention, mixed FP16/INT8 storage, and a pyramidal per-layer budget variant. Across four model families and generated lengths up to 4K, \ourmethodq\ stays near the footprint of a fixed 512-token sliding window while remaining within 1.5--2.1 perplexity points of full KV. On Needle-in-a-Haystack up to 32K tokens, \ourmethod reaches 91.4\% retrieval accuracy versus 53.8\% for sliding windows and 80.6\% for \textsc{H2O}; on 75 VisualWebArena tasks it retains 95.3\% of full-KV success at 2.8$\times$ lower peak memory.
\end{abstract}

\section{Introduction}

Long-horizon LLM applications such as web agents, long-document analysis, multi-turn assistants, and tool-using systems accumulate context over many interaction rounds. The KV cache grows linearly with sequence length and depth; in our full-KV Qwen-32B measurement for the 4K generated-token sweep, which retains the matched prefill/prefix state, the measured KV-memory allocator footprint reaches 15.8 GB. This cache can dominate both GPU memory and attention latency, so cache management has become a first-order systems problem rather than an implementation detail.

Many cache policies answer the same question with signals from the past. Sliding-window attention keeps recent tokens; \textsc{H2O} keeps historical heavy hitters; \textsc{Scissorhands} uses persistence of importance; \textsc{SnapKV} decides from prompt-phase observations; and \textsc{PyramidKV} varies the budget by layer~\citep{streamingllm,h2o,scissorhands,snapkv,pyramidkv}. Recent adaptive methods improve the allocation of head, layer, or precision budgets, but they still leave open a complementary question: can the current output distribution tell the cache manager \emph{when} a decoding step needs more retained context? The next-token distribution is already available before sampling, and its shape gives a direct measurement of how uncertain the model is at the current step. The resulting budget expansion is forward-looking: it cannot recover tokens already evicted, but it can prevent premature eviction while the model processes difficult spans, while the attention--recency ranker keeps older high-utility tokens alive.

This paper asks whether that free signal can improve the memory--quality trade-off of KV-cache eviction. We answer with \ourmethod, a confidence-aware cache manager for autoregressive generation. At each step, \ourmethod maps entropy, log-probability margin, and top-token mass into a bounded confidence score. The score selects a tight or loose cache budget. Within the selected budget, the manager evicts low-ranked tokens according to a combination of exponential-moving-average attention mass and recency, subject to a hard protected window over the newest tokens. The design is deliberately simple: it does not require training, does not change model weights, and can be inserted into a standard generation loop.

The key empirical result is that confidence controls \emph{when} to evict more effectively than static policies. At matched memory, \ourmethod-L closes 74\% of the perplexity gap between a 512-token sliding window and full KV on GPT-2; \ourmethodq closes 60\%. Unless otherwise stated, \ourmethod-L denotes the pyramidal layer-budget variant with the same FP16/INT8 storage used by \ourmethodq. The same policy preserves long-range retrieval in Needle-in-a-Haystack, preserves web-agent success on VisualWebArena, and improves throughput in completed batch-size comparisons. A matched-rate isolation study shows that the improvement is not merely a consequence of evicting less often: random eviction with the same schedule degrades to 36.54 perplexity, while full \ourmethod is 30.92.

Our contributions are:
\begin{itemize}[leftmargin=1.3em,itemsep=1pt,topsep=2pt]
  \item a confidence-aware KV eviction policy that uses the current next-token distribution to choose a per-step cache budget;
  \item a systems design that combines adaptive eviction with blockwise attention, cache compaction, mixed FP16/INT8 KV storage, and an optional pyramidal per-layer budget;
  \item a mechanistic test showing that confidence anticorrelates with the KL shift induced by ablating recent context, supporting the policy's central assumption;
  \item a matched-memory evaluation against sliding windows, \textsc{H2O}, \textsc{Scissorhands}, \textsc{SnapKV}, and \textsc{PyramidKV} across language-model, long-context retrieval, web-agent, latency, and throughput workloads.
\end{itemize}

\section{Related work}

\paragraph{KV-cache eviction.}
Sliding windows and attention sinks give robust streaming behavior with fixed memory, but they discard old tokens regardless of semantic utility~\citep{streamingllm,longformer}. \textsc{H2O} and \textsc{Scissorhands} rank cached tokens by accumulated or persistent attention importance~\citep{h2o,scissorhands}. \textsc{SnapKV} and \textsc{FastGen} infer useful cache structure during the prompt or by head-wise policy selection~\citep{snapkv,fastgen}. \textsc{PyramidKV} observes that deeper layers can often use smaller budgets~\citep{pyramidkv}. \ourmethod is complementary to these lines: its distinctive signal is step-level uncertainty from the current output distribution, and its budget adapts over time rather than being fixed by a global cap or prompt-phase decision.

\paragraph{Adaptive and uncertainty-aware KV compression.}
A growing line of work adapts the compression budget rather than using one global cap. \textsc{Ada-KV} derives a loss-guided view of eviction and allocates budgets across attention heads~\citep{adakv}; \textsc{ZigZagKV} uses layer uncertainty to allocate layer-specific budgets~\citep{zigzagkv}; and \textsc{UNComp} uses matrix-entropy uncertainty to expose sparsity and long-range retrieval structure during compression~\citep{uncomp}. Adaptive precision methods similarly choose bit-width from token-level features, including entropy-based uncertainty~\citep{adaptivekvquant}. These works are closest in spirit because they make the cache policy data-dependent. \ourmethod differs in the control signal and axis of adaptation: it reads the target model's current next-token distribution and uses that signal to choose a decoding-time token-retention budget, then validates the assumed confidence/context-demand relation with a KL-ablation test. This distinction is useful in practice because it composes with head-wise allocation, layer-wise allocation, and precision selection rather than replacing them.

\paragraph{Efficient attention and serving memory.}
\textsc{FlashAttention} reduces activation memory using tiled online softmax, but it does not decide which KV entries should remain cached~\citep{flashattention,flashattention2}. \textsc{PagedAttention} improves serving layout and batching by paging KV blocks~\citep{pagedattention}; \textsc{Quest} and \textsc{SparQ} sparsify attention reads~\citep{quest,sparq}. Quantization systems such as \textsc{KIVI} and \textsc{KVQuant} reduce cache precision without changing token retention~\citep{kivi,kvquant}. \ourmethod composes with these techniques because it addresses the orthogonal question of which tokens live in the cache.

\paragraph{Confidence signals.}
Confidence has been used for early exit and speculative decoding, where it controls computation or acceptance decisions~\citep{earlyexit,specinfer,medusa}. \ourmethod uses the same kind of signal to control memory state. Our novelty claim is therefore narrow: not that uncertainty has never been used for compression, but that next-token confidence can drive a per-step KV token-retention budget during autoregressive decoding.

\begin{table}[H]
\centering
\caption{KV-cache management methods differ in the signal used to decide token utility and in whether the budget adapts. \ourmethod is distinctive because the budget reacts to the current next-token distribution.}
\label{tab:taxonomy}
\small
\setlength{\tabcolsep}{3pt}
\begin{tabular}{@{}p{0.22\linewidth}p{0.24\linewidth}p{0.19\linewidth}p{0.17\linewidth}p{0.10\linewidth}@{}}
\toprule
\textbf{Method} & \textbf{Signal} & \textbf{Budget axis} & \textbf{Granularity} & \textbf{Quant.} \\
\midrule
Sliding window & recency & fixed & token & no \\
\textsc{H2O}/\textsc{Scissorhands} & historical attention & fixed cap & head/token & no \\
\textsc{SnapKV}/\textsc{FastGen} & prompt/head obs. & prompt/head & head/token & no \\
\makecell[l]{\textsc{PyramidKV}/\\\textsc{ZigZagKV}} & attn/layer uncertainty & layer & layer/token & no \\
\textsc{Ada-KV} & attention-output loss & head & head/token & no \\
\textsc{KIVI}/\textsc{KVQuant} & none (precision only) & none & storage & yes \\
Adaptive precision & token features, entropy & bit-width & token/storage & yes \\
\ourmethod(+INT8) & next-token confidence & decoding step & token/layer & optional \\
\bottomrule
\end{tabular}
\end{table}

\section{Method}
\label{sec:method}

\begin{figure}[H]
\centering
\includegraphics[width=1\linewidth]{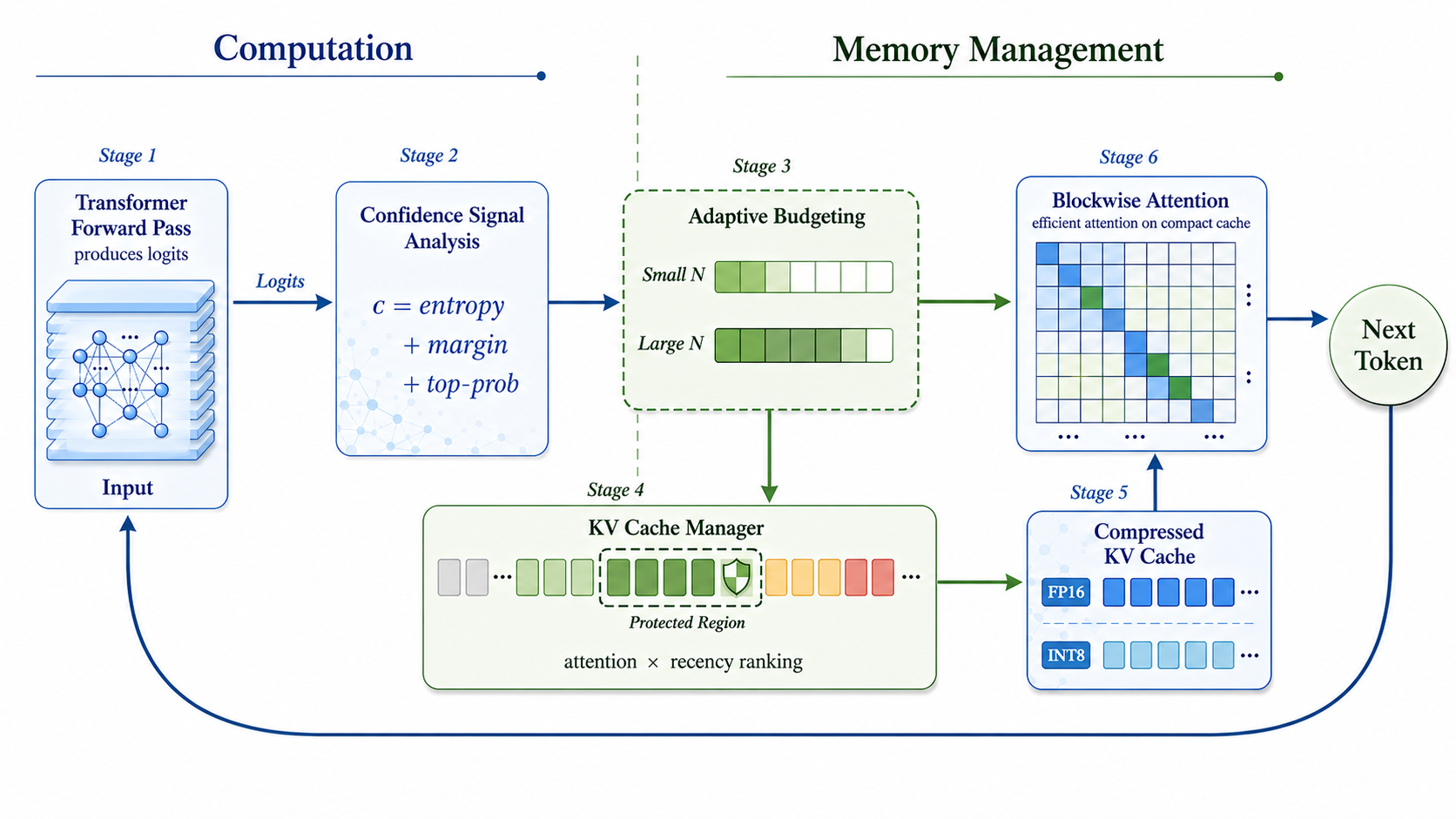}
\caption{\ourmethod pipeline. The logits at each decoding step are converted into confidence $c$, which selects a tight or loose cache budget. The cache manager ranks candidates by attention mass and recency, protects the most recent tokens, compacts the survivors, and serves the next attention call from a mixed-precision cache.}
\label{fig:system}
\end{figure}

\paragraph{Cache manager.}
Each layer owns pre-allocated K/V tensors, a valid-length counter, and parallel metadata storing original position, generation step, and an exponential moving average (EMA) of attention mass. Appends are $O(1)$. When eviction is triggered, the manager gathers surviving entries into contiguous storage. This contiguous layout keeps the attention kernel simple and avoids adding an indirection table to every read.

\paragraph{Confidence estimator.}
Let $\ell$ be the logits and $p=\mathrm{softmax}(\ell)$ the next-token distribution. We compute normalized entropy $\hat{H}=H(p)/\log V$, log-probability margin $m=\log p_{(1)}-\log p_{(2)}$, and top-token probability $p_{(1)}$; $\sigma(\cdot)$ denotes the logistic sigmoid. The confidence score is
\begin{equation}
c = w_H(1-\hat{H}) + w_m \sigma(m) + w_p p_{(1)}, \qquad (w_H,w_m,w_p)=(0.4,0.3,0.3).
\label{eq:confidence}
\end{equation}
The weights were chosen by a small grid search on GPT-2 and were stable across the evaluated models. The score need not be calibrated as a probability; the policy only requires a monotone relation between confidence and context demand.

\paragraph{Budget and ranker.}
Given threshold $\tau$, \ourmethod chooses
\[
N = \begin{cases}
N_\mathrm{high}, & c \geq \tau,\\
N_\mathrm{low}, & c < \tau,
\end{cases}
\qquad N_\mathrm{high}\leq N_\mathrm{low}.
\]
Here $N_\mathrm{high}$ is the tighter budget used on confident steps, and $N_\mathrm{low}$ is the larger budget used on uncertain steps. For a candidate token $i$, the ranker computes
\[
s_i = \alpha \hat{a}_i + (1-\alpha)\hat{r}_i,
\]
where $\hat{a}_i$ is normalized EMA attention mass and $\hat{r}_i$ is normalized recency. For token $i$ in layer $\ell$, attention mass is averaged over heads and then updated as $a_i\leftarrow \lambda a_i+(1-\lambda)\bar A_i$ with $\lambda=0.9$. Both attention mass and recency are min--max normalized over the non-protected candidates in the current layer before interpolation; recency uses original generation step, so newer retained tokens receive larger $\hat r_i$. The manager evicts the lowest-scored entries until the cache length is at most $N$, while always retaining the most recent $P$ tokens. The protected window prevents pathological deletion of tokens still in the local attention working set.

\begin{algorithm}[H]
\caption{\ourmethod one-step generation}
\label{alg:confkv}
\begin{algorithmic}[1]
\REQUIRE cache $\mathcal{C}$; model $f$; input token $x_t$; threshold $\tau$; budgets $N_\mathrm{high},N_\mathrm{low}$; protected window $P$; ranker weight $\alpha$; FP16 window $W$
\STATE $\mathrm{logits}, A \leftarrow f(x_t,\mathcal{C})$
\STATE $p \leftarrow \mathrm{softmax}(\mathrm{logits})$
\STATE $c \leftarrow w_H(1-\hat{H}(p)) + w_m\sigma(\log p_{(1)}-\log p_{(2)}) + w_p p_{(1)}$
\STATE $N \leftarrow N_\mathrm{high}$ if $c\geq \tau$ else $N_\mathrm{low}$
\FOR{each layer $\ell$}
  \STATE update EMA attention metadata using $A^{(\ell)}$
  \IF{$|\mathcal{C}_\ell|>N$}
    \STATE rank non-protected tokens by $s_i=\alpha\hat{a}_i+(1-\alpha)\hat{r}_i$
    \STATE evict $|\mathcal{C}_\ell|-N$ lowest-ranked tokens and compact survivors
  \ENDIF
\STATE quantize retained positions outside the most recent $W$ generated tokens to INT8
\ENDFOR
\STATE append new K/V in FP16 and sample $x_{t+1}$ from $p$
\end{algorithmic}
\end{algorithm}

\paragraph{Tiled attention and mixed precision.}
Attention reads the compacted cache in blocks and maintains the running max and normalizer needed for exact online softmax. The most recent $W$ retained tokens by original generation step remain in FP16; older retained entries are symmetrically quantized to INT8 per head and channel with scale $s=\max(|x|)/127$. Dequantization is fused into the blockwise attention read. This mixed representation gives most of the memory benefit of lower precision while avoiding the larger perplexity loss observed with NF4 and INT4 in our ablations.

\paragraph{Pyramidal layer budgets.}
\ourmethod-L allocates the selected budget non-uniformly across layers and uses the same FP16/INT8 storage path as \ourmethodq:
\[
N^{(\ell)}_\mathrm{high}=\max(N_{\min},N^{(0)}_\mathrm{high}\beta^{\ell/L}),
\]
with an analogous expression for $N_\mathrm{low}$, where $L$ is the number of layers. We use $\beta=0.5$ and $N_{\min}=96$ unless noted. This follows the observation that deeper layers often concentrate useful information into fewer tokens~\citep{pyramidkv}. The mechanism adapts along depth, while the confidence rule adapts over time.

\section{Experimental setup}
\label{sec:setup}

\paragraph{Models and workloads.}
We evaluate GPT-2 (124M), Qwen-14B, gpt-oss-20b, and Qwen-32B~\citep{gpt2,qwen,gptoss}. WikiText-2 continuation perplexity is measured at 512, 1024, 2048, and 4096 generated tokens with a matched prefill/prefix length retained in the KV cache for memory measurements~\citep{wikitext}. Needle-in-a-Haystack (NIAH) uses haystack lengths from 1K to 32K and five needle depths~\citep{niah}. VisualWebArena (VWA) uses gpt-oss-20b with no raw image tokens: our wrapper serializes each rendered page into visible text, OCR text, element identifiers, bounding boxes, and the previous action, then emits the standard VWA text action. We evaluate 75 tasks stratified across shopping, navigation, form, and information-seeking categories, up to 30 steps, and 256 tokens per step~\citep{visualwebarena}. Throughput sweeps batch sizes from 1 to 32 at a 2048-token generation length.

\paragraph{Baselines and configurations.}
Baselines are full KV, a fixed 512-token sliding window, \textsc{H2O}, \textsc{Scissorhands}, \textsc{SnapKV}, and \textsc{PyramidKV}. We tune baselines to match \ourmethodq's average peak KV memory when reporting head-to-head results and run them with the same tiled attention path to isolate the cache policy. Unless otherwise stated, $\tau=0.7$, $N_\mathrm{high}=128$ for WikiText and 256 for NIAH/VWA, $N_\mathrm{low}=256$ for WikiText and 512 for NIAH/VWA, $P=32$ for WikiText and 64 for NIAH/VWA, $\alpha=0.65$, FP16 window $W=128$ or 256, and block size $B=128$.

\paragraph{Statistics and compute.}
WikiText-2 is deterministic under greedy decoding and is reported once. NIAH, VWA, and throughput are reported as mean $\pm$ standard deviation over three seeds. Experiments run on NVIDIA H100 80 GB GPUs with CUDA 12.8, PyTorch 2.9, and Transformers 4.51; additional hardware and hyperparameter details appear in Appendix~\ref{app:repro}.

\section{Results}
\label{sec:results}

\begin{table}[H]
\centering
\caption{Matched-memory comparison on GPT-2 at 2048 generated tokens. PPL is WikiText-2 continuation; latency is per-step. Baselines are calibrated to match \ourmethodq's average peak KV memory (38.7 MB $\pm$ 2 MB).}
\label{tab:head2head}
\small
\setlength{\tabcolsep}{3.5pt}
\begin{tabular}{@{}lrrrr@{}}
\toprule
\textbf{Method} & \textbf{Mem. MB} & \textbf{PPL} & \textbf{p50 ms} & \textbf{p95 ms} \\
\midrule
Full KV & 151.0 & 29.14 & 4.65 & 7.91 \\
Sliding-512 & 38.6 & 34.37 & 2.08 & 2.47 \\
\textsc{H2O} & 39.1 & 31.78 & 2.41 & 3.52 \\
\textsc{Scissorhands} & 39.2 & 31.94 & 2.43 & 3.58 \\
\textsc{SnapKV} & 42.4 & 31.41 & 2.28 & 3.26 \\
\textsc{PyramidKV} & 37.8 & 31.09 & 2.46 & 3.71 \\
\ourmethod & 52.8 & 30.92 & 2.51 & 4.12 \\
\ourmethodq & 38.7 & 31.26 & 2.64 & 4.35 \\
\ourmethod-L & \textbf{34.2} & \textbf{30.48} & 2.57 & 4.18 \\
\bottomrule
\end{tabular}
\end{table}

\paragraph{Matched-memory quality.}
Table~\ref{tab:head2head} gives the central comparison. At the same memory scale as the sliding window, \ourmethodq improves perplexity by 3.11 points, while \ourmethod-L improves by 3.89 points and uses less memory than every baseline. Relative to the full-cache/sliding-window gap, \ourmethod-L closes 74\% of lost quality. The best static or historical-attention baseline, \textsc{PyramidKV}, closes 63\% of the gap. This difference matters most when only a few steps require extra context: a static cap must either over-provision all steps or under-provision the rare difficult ones.

\begin{figure}[H]
\centering
\begin{minipage}{0.49\linewidth}
\centering
\includegraphics[width=\linewidth]{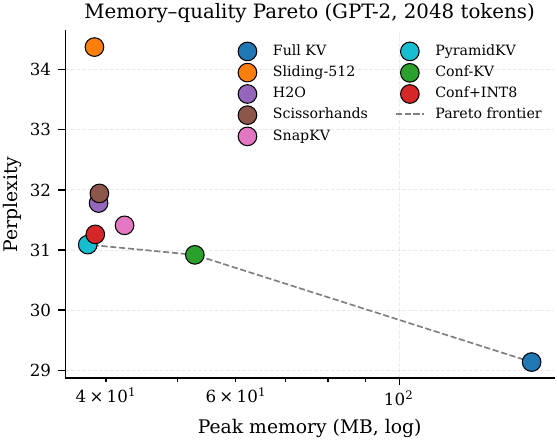}
\end{minipage}
\hfill
\begin{minipage}{0.49\linewidth}
\centering
\includegraphics[width=\linewidth]{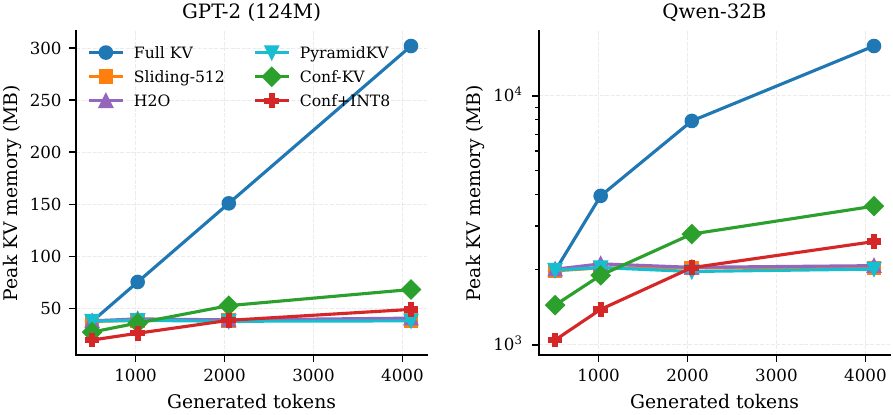}
\end{minipage}
\caption{Memory--quality and scaling behavior. Left: GPT-2 memory--PPL Pareto at 2048 tokens. Right: peak KV memory across generated-token lengths for GPT-2 and Qwen-32B.}
\label{fig:memory-quality}
\end{figure}

Figure~\ref{fig:memory-quality} shows that the \ourmethod variants lie on the Pareto frontier. Across all four evaluated models, \ourmethodq keeps peak memory in the same range as a 512-token sliding window, from parity at the main 2048-token comparison to about 1.3$\times$ the sliding footprint at the longest sweep, while retaining much more quality. On Qwen-32B at 4K generated tokens with matched prefill state, the absolute KV-memory reduction is 13.2 GB (15.8 GB to 2.6 GB), which changes feasible batch size on a single 80 GB H100.

\paragraph{Does confidence itself matter?}
We isolate the confidence signal with matched-rate baselines: each variant uses the same per-step eviction probability and the same number of evicted tokens per event as \ourmethod, but changes which tokens are removed or how they are ranked. Random eviction at the same rate gives 36.54 PPL, worse than the fixed sliding window. Recency-only ranking gives 32.08, attention-only gives 31.47, and full \ourmethod gives 30.92. Thus the ranker and the confidence-gated schedule both contribute. We further test the policy's premise by ablating the past 256 tokens in a 10K-step GPT-2 trace and measuring the KL shift in the next-token distribution. Confidence and KL shift have Pearson $r=-0.77$ ($p<10^{-30}$), with a monotone decreasing binned mean. The effect repeats on Qwen-14B, gpt-oss-20b, and Qwen-32B (Appendix~\ref{app:calibration}).

\begin{figure}[H]
\centering
\begin{minipage}{0.49\linewidth}
\centering
\includegraphics[width=\linewidth]{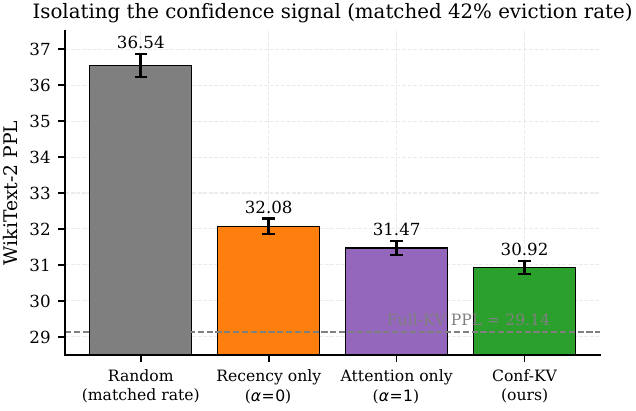}
\end{minipage}
\hfill
\begin{minipage}{0.49\linewidth}
\centering
\includegraphics[width=\linewidth]{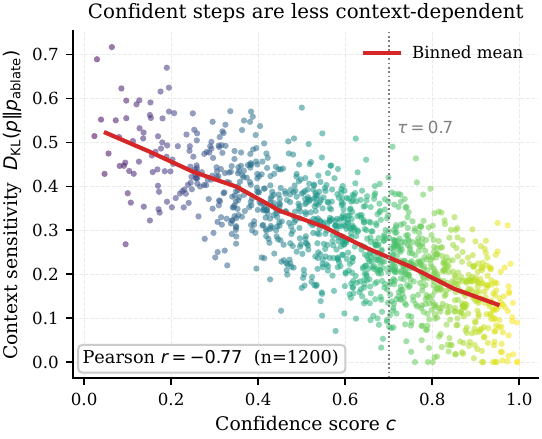}
\end{minipage}
\caption{Signal validation. Left: matched-rate isolation shows that random eviction with the same schedule fails, while attention and recency each help. Right: high confidence anticorrelates with the KL shift caused by ablating recent context.}
\label{fig:signal}
\end{figure}

\begin{figure}[H]
\centering
\includegraphics[width=\linewidth]{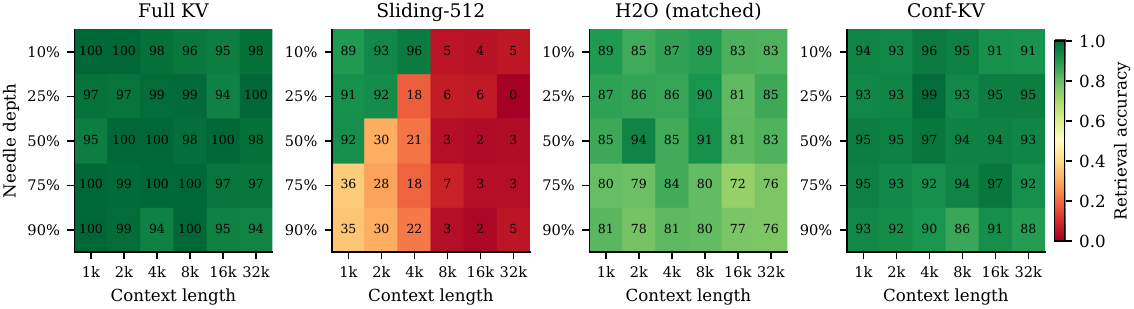}
\caption{Needle-in-a-Haystack retrieval accuracy on gpt-oss-20b. Columns are haystack lengths; rows use distance-from-query depth, so larger percentages place the needle farther from the retained recent window. \ourmethod degrades more gracefully than fixed-window and historical-attention baselines.}
\label{fig:niah}
\end{figure}

\paragraph{Long-context retrieval.}
NIAH stresses whether an eviction policy can retain an old but decisive fact. Figure~\ref{fig:niah} reports depth as distance from the query/end of the prompt: 10\% is near the retained recent window, while 90\% is older context. The sliding window therefore fails structurally once the needle lies outside the last 512 tokens. \textsc{H2O} performs better but loses middle-depth needles in long haystacks because rare tokens may not accumulate attention until the query arrives. \ourmethod reaches 91.4\% average accuracy across the grid versus 53.8\% for sliding and 80.6\% for \textsc{H2O}. The trace-level behavior matches the design: confidence drops during the retrieval query, which immediately expands the budget. The remaining failures occur when the model remains over-confident before rare-entity lookup; raising $\tau$ from 0.7 to 0.8 recovers most of these cases at a 12\% memory cost.

\paragraph{VisualWebArena.}
On 75 VWA tasks, \ourmethod retains 95.3\% of full-KV task success while reducing peak memory by 2.8$\times$. The full-KV success rate is 40.2\%, and \ourmethod reaches 38.3\%; the gap is within one standard deviation for each category. Sliding-window truncation loses 11.1 absolute points. The largest gains over \textsc{H2O} appear in information-seeking tasks, where page re-reading causes confidence dips that give \ourmethod extra budget exactly when the agent needs to recover older observations.

\begin{figure}[H]
\centering
\begin{minipage}{0.49\linewidth}
\centering
\includegraphics[width=\linewidth]{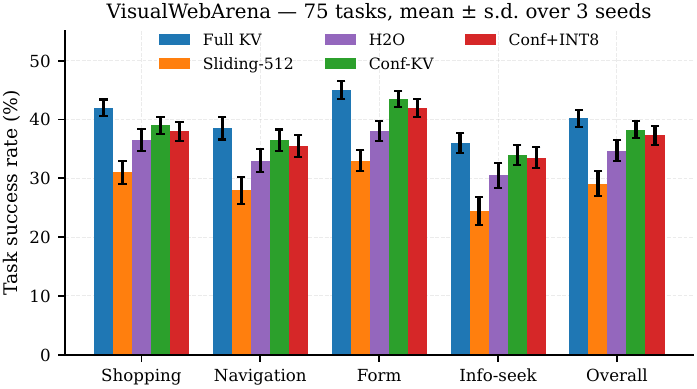}
\end{minipage}
\hfill
\begin{minipage}{0.49\linewidth}
\centering
\includegraphics[width=\linewidth]{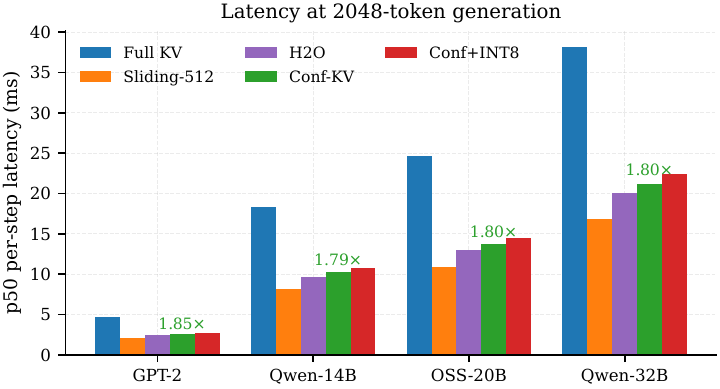}
\end{minipage}
\caption{Task quality and latency. Left: VWA success rate, mean $\pm$ s.d. over three seeds. Right: p50 per-step latency at 2048 generated tokens across four models.}
\label{fig:vwa-latency}
\end{figure}

\paragraph{Latency, profiling, and batching.}
At 2048 tokens, \ourmethod reduces p50 latency by 1.8$\times$ on GPT-2 and 1.8$\times$ on Qwen-32B relative to full KV, while \ourmethodq is slightly slower than \ourmethod because of quant/dequant overhead. Profiling shows the trade-off explicitly: attention falls from 62\% of full-KV step time to 47\% under \ourmethod, while compaction adds 0.22 ms and metadata updates add 0.11 ms on GPT-2. Throughput benefits grow with batch size because memory is the bottleneck. In our prototype allocator, the full-KV throughput run did not complete at batch 16 despite GPT-2's small theoretical KV footprint; we therefore rely on the completed batch-$\leq$8 comparisons for speedup claims. At batch 8, \ourmethod gives 2.06$\times$ the full-KV throughput while matching the sliding-window baseline within 1\%.

\begin{figure}[H]
\centering
\begin{minipage}{0.49\linewidth}
\centering
\includegraphics[width=\linewidth]{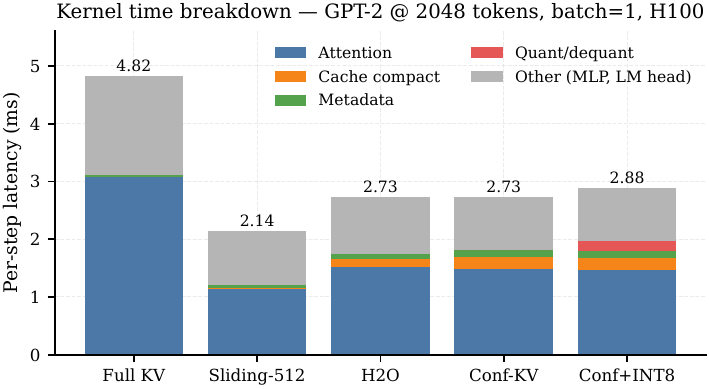}
\end{minipage}
\hfill
\begin{minipage}{0.49\linewidth}
\centering
\includegraphics[width=\linewidth]{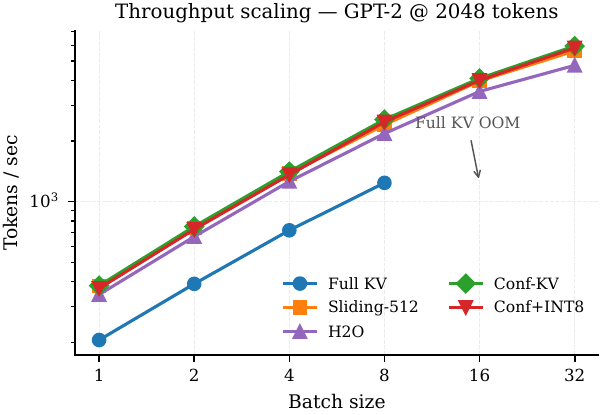}
\end{minipage}
\caption{Systems behavior. Left: per-step latency decomposition on GPT-2. Right: GPT-2 throughput versus batch size at 2048 generated tokens; full-KV points beyond batch 8 were unavailable in our prototype allocator run and are not used for speedup claims.}
\label{fig:systems}
\end{figure}

\section{Ablations and sensitivity}
\label{sec:ablations}

\begin{figure}[H]
\centering
\begin{minipage}{0.49\linewidth}
\centering
\includegraphics[width=\linewidth]{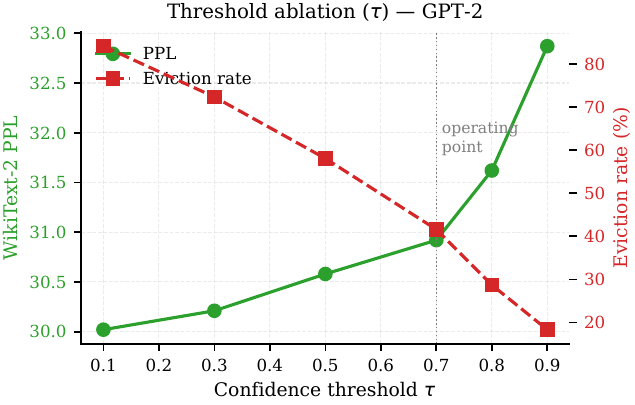}
\end{minipage}
\hfill
\begin{minipage}{0.49\linewidth}
\centering
\includegraphics[width=\linewidth]{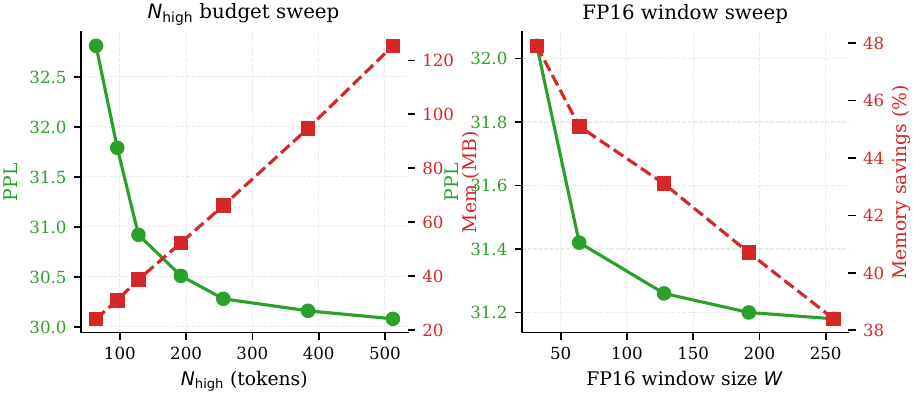}
\end{minipage}
\caption{Ablations. Left: $\tau$ controls the eviction-rate/quality trade-off. Right: budget and FP16-window sweeps place the default $N_\mathrm{high}=128$ and $W=128$ near the knees of the curves.}
\label{fig:ablation-main}
\end{figure}

\paragraph{Threshold and budget.}
The confidence threshold $\tau$ controls how often the policy takes the tight budget. Figure~\ref{fig:ablation-main} shows that $\tau=0.7$ is a stable operating point: lower thresholds make $c\geq\tau$ easier to satisfy, increase eviction, and reduce memory at a quality cost; higher thresholds are more conservative and improve quality at a memory cost. The $N_\mathrm{high}$ sweep shows a similar knee at 128 tokens for WikiText-2. Below this point, confident-step evictions are too aggressive; above it, memory rises faster than quality improves. These curves are useful operationally because they expose a direct quality--memory knob rather than hiding the trade-off inside a learned policy.

\paragraph{Precision and layer allocation.}
The FP16 window $W$ controls how much of the recent cache avoids quantization. $W=128$ is the best GPT-2 operating point in our sweep: smaller windows expose recently generated tokens to quantization error, while larger windows reduce memory savings. INT8 per-(head,channel) scaling gives 0.38\% mean roundtrip error and adds only 0.34 PPL relative to FP16 cache storage; NF4 and INT4 add 0.91 and 1.65 PPL beyond INT8, respectively. Table~\ref{tab:head2head} reports the pyramidal \ourmethod-L result; it improves GPT-2 PPL by 0.44 over uniform \ourmethod while reducing measured peak memory. Appendix~\ref{app:additional} gives the full quantization and live-memory traces.

\begin{figure}[H]
\centering
\begin{minipage}{0.49\linewidth}
\centering
\includegraphics[width=\linewidth]{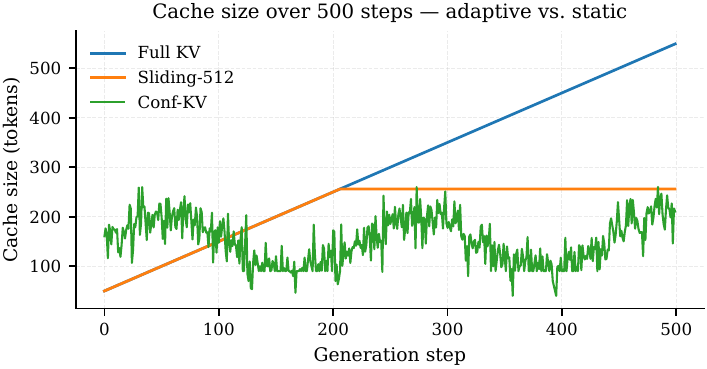}
\end{minipage}
\hfill
\begin{minipage}{0.49\linewidth}
\centering
\includegraphics[width=\linewidth]{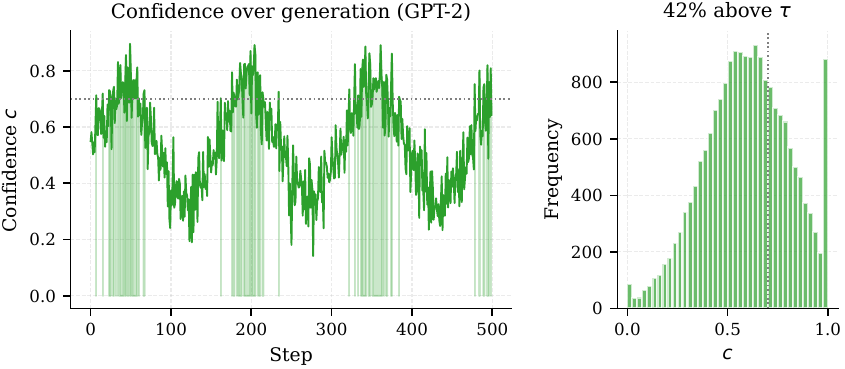}
\end{minipage}
\caption{Trace-level behavior. Left: cache size over 500 generation steps. Right: confidence trace and empirical histogram. The policy's sawtooth envelope follows confidence rather than a fixed window.}
\label{fig:trace-main}
\end{figure}

\paragraph{Temporal behavior.}
\ourmethod does not maintain a constant cache size. Figure~\ref{fig:trace-main} shows a sawtooth trajectory: the cache grows during uncertain or below-threshold steps, then compacts when confidence rises and a tight budget is selected. The confidence histogram is bimodal enough that the policy receives a steady supply of high-confidence opportunities, but it still preserves a heavy tail of low-confidence steps. This trace-level behavior explains why fixed-memory comparisons alone understate the advantage of adaptivity: the average memory can match a static baseline while the per-step memory is spent where it matters.

\section{Implementation notes and failure modes}
\label{sec:implementation}

\paragraph{Compaction cost and memory envelope.}
The cache manager uses contiguous compaction rather than a paged indirection table. This choice keeps the attention kernel identical to a dense tiled read, but it pays a gather cost on eviction events. In our GPT-2 profile, compaction is 0.22 ms per step at the observed eviction rate and metadata updates are 0.11 ms, together smaller than the attention savings from the reduced cache. The live-memory trace in Appendix~\ref{app:additional} shows bounded post-compaction behavior; Appendix~\ref{app:extended} reports measured allocator footprints, which also include prefix/prompt storage and temporary workspaces.

\paragraph{Observed failure modes.}
The remaining NIAH failures occur when the model is confidently wrong before a rare-entity lookup, so the policy selects the tight budget when it should preserve context. Raising $\tau$ recovers these examples at a measurable memory cost, which suggests that the trade-off is tunable rather than arbitrary. The VWA failures have the same structure: product names, form fields, or rare entities observed hundreds of tokens earlier can be evicted during a high-confidence stretch. In contrast, sliding-window failures are usually structural because the relevant token is outside the fixed window regardless of the model state.

\paragraph{Compatibility with serving systems.}
\ourmethod decides which tokens remain live; it does not require a specific physical layout. A production implementation could replace contiguous compaction with a block table, but token-level eviction would create partially dead blocks in a \textsc{PagedAttention}-style layout unless eviction is coarsened or sparse masks are added. The method is also orthogonal to speculative decoding: a draft model can propose candidate tokens while the target model's accepted-step logits continue to drive the cache budget.

\section{Limitations and conclusion}
\label{sec:limitations}

\ourmethod is a no-op on short contexts that never exceed the eviction threshold, and its speedups are smaller when MLP compute rather than attention or KV bandwidth dominates latency. The confidence signal also becomes less informative under high-temperature sampling, where entropy saturates and $\tau$ may need retuning. Finally, the contiguous compaction strategy is simple and fast enough in our setting, but mapping token-level eviction to a paged serving layout is not free: fine-grained eviction creates partially dead blocks unless the implementation coarsens eviction or adds sparse masks.

Lower KV-cache memory can reduce inference cost and energy use for long-context systems, but it can also lower the cost of undesirable long-horizon automation. \ourmethod does not add new model capabilities; deployments should inherit the safety controls used for the underlying agent or LLM. Overall, the results show that current uncertainty is useful systems metadata for improving the memory--quality Pareto while remaining compatible with paging, quantization, and speculative decoding.

\clearpage
\bibliographystyle{plainnat}
\bibliography{references}

\clearpage
\appendix

\section{Extended quantitative results}
\label{app:extended}

\begin{table}[H]
\centering
\caption{Measured peak KV-memory allocator footprint (MB) across all models and generated-token counts. The continuation sweep keeps a matched prefill/prefix state in the cache, and measurements include temporary cache-management workspaces, so values are not identical to the live post-compaction token count.}
\label{tab:full_memory}
\small
\setlength{\tabcolsep}{3pt}
\begin{tabular}{@{}llrrrr@{}}
\toprule
\textbf{Model} & \textbf{Config} & \textbf{Gen. 512} & \textbf{Gen. 1024} & \textbf{Gen. 2048} & \textbf{Gen. 4096} \\
\midrule
GPT-2 & Full KV & 37.7 & 75.5 & 151.0 & 302.0 \\
 & Sliding-512 & 37.7 & 38.6 & 38.6 & 38.6 \\
 & \ourmethodq & 19.8 & 26.3 & 38.7 & 49.2 \\
 & \ourmethod-L & 18.3 & 24.1 & 34.2 & 42.6 \\
\midrule
Qwen-14B & Full KV & 890 & 1780 & 3560 & 7120 \\
 & Sliding-512 & 890 & 915 & 915 & 915 \\
 & \ourmethodq & 469 & 623 & 917 & 1167 \\
 & \ourmethod-L & 438 & 572 & 840 & 1082 \\
\midrule
gpt-oss-20b & Full KV & 1240 & 2480 & 4960 & 9920 \\
 & Sliding-512 & 1240 & 1275 & 1275 & 1275 \\
 & \ourmethodq & 653 & 868 & 1277 & 1625 \\
 & \ourmethod-L & 611 & 798 & 1173 & 1512 \\
\midrule
Qwen-32B & Full KV & 1980 & 3960 & 7920 & 15840 \\
 & Sliding-512 & 1980 & 2035 & 2035 & 2035 \\
 & \ourmethodq & 1044 & 1386 & 2039 & 2595 \\
 & \ourmethod-L & 978 & 1276 & 1871 & 2408 \\
\bottomrule
\end{tabular}
\end{table}

\section{Additional ablations and traces}
\label{app:additional}

\begin{figure}[H]
\centering
\includegraphics[width=0.78\linewidth]{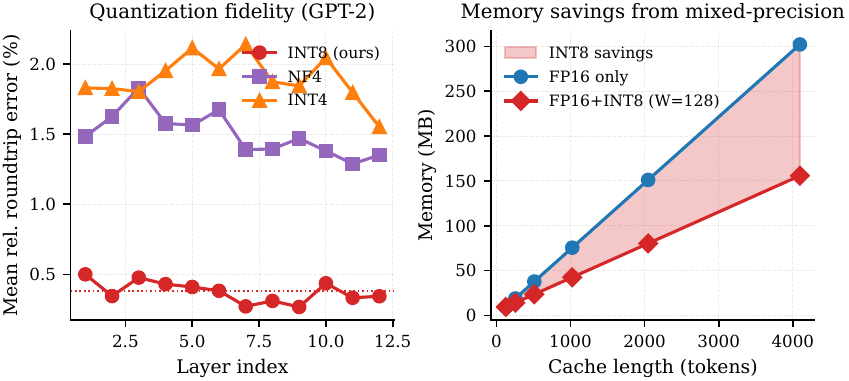}
\caption{Quantization analysis. INT8 gives most of the cache-memory reduction with lower roundtrip error than NF4 or INT4.}
\label{fig:precision-pyramid}
\end{figure}

\begin{figure}[H]
\centering
\includegraphics[width=0.62\linewidth]{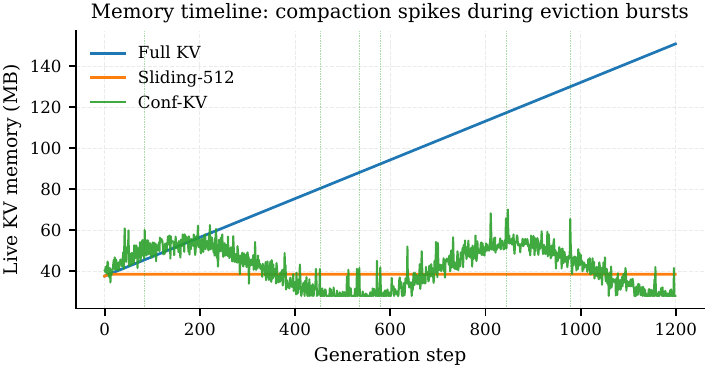}
\caption{Live KV memory over 1200 generation steps; adaptive compaction yields bounded sawtooth behavior in the traced decode.}
\label{fig:memory-timeline}
\end{figure}

\section{Confidence calibration across models}
\label{app:calibration}

\begin{table}[H]
\centering
\caption{Pearson correlation between confidence score and KL shift from ablating the past 256 tokens.}
\label{tab:calib}
\small
\begin{tabular}{@{}lrr@{}}
\toprule
\textbf{Model} & \textbf{$n$} & \textbf{Pearson $r$} \\
\midrule
GPT-2 & 1{,}200 & $-0.77$ \\
Qwen-14B & 1{,}200 & $-0.38$ \\
gpt-oss-20b & 1{,}200 & $-0.41$ \\
Qwen-32B & 1{,}200 & $-0.36$ \\
\bottomrule
\end{tabular}
\end{table}

\section{Reproducibility details}
\label{app:repro}

All experiments use a single NVIDIA H100 80 GB GPU unless otherwise noted. VWA evaluation uses the official VisualWebArena environment hosted on AWS EC2. The implementation wraps HuggingFace \texttt{CausalLM} models through a confidence-aware generator and keeps the cache manager, quantizer, and tiled attention backend as separable modules.

\begin{table}[H]
\centering
\caption{Hyperparameter settings.}
\label{tab:hparams}
\small
\begin{tabular}{@{}llll@{}}
\toprule
 & \textbf{WikiText-2} & \textbf{NIAH} & \textbf{VWA} \\
\midrule
$\tau$ & 0.7 & 0.7 & 0.7 \\
$N_\mathrm{high}$ & 128 & 256 & 256 \\
$N_\mathrm{low}$ & 256 & 512 & 512 \\
$P$ & 32 & 64 & 64 \\
$\alpha$ & 0.65 & 0.70 & 0.65 \\
EMA decay $\lambda$ & 0.90 & 0.90 & 0.90 \\
FP16 window $W$ & 128 & 256 & 256 \\
Block size $B$ & 128 & 128 & 128 \\
\ourmethod-L $(\beta,N_{\min})$ & (0.5, 96) & (0.5, 96) & (0.5, 96) \\
Seeds & 1 & 3 & 3 \\
\bottomrule
\end{tabular}
\end{table}

\begin{table}[H]
\centering
\caption{Existing assets, versions, and license/terms status. We cite the original sources and do not redistribute model weights, datasets, benchmark environments, or baseline code.}
\label{tab:assets}
\small
\setlength{\tabcolsep}{3pt}
\begin{tabular}{@{}p{0.22\linewidth}p{0.23\linewidth}p{0.45\linewidth}@{}}
\toprule
\textbf{Asset} & \textbf{Use} & \textbf{License / terms} \\
\midrule
GPT-2 & language-model baseline & Modified MIT license on OpenAI GPT-2 release. \\
Qwen-14B/32B & scale-out language models & Qwen model-card terms for the public checkpoints used in evaluation. \\
gpt-oss-20b & NIAH and VWA model & Apache 2.0 license plus OpenAI gpt-oss usage policy. \\
WikiText-2 & perplexity benchmark & Creative Commons Attribution-ShareAlike/GFDL terms for WikiText. \\
Needle-in-a-Haystack & retrieval benchmark & MIT license in the benchmark repository. \\
VisualWebArena & web-agent benchmark & MIT license for the benchmark code and task environment. \\
Baselines & comparative methods & Reimplemented from cited papers; no third-party baseline code is redistributed. \\
\bottomrule
\end{tabular}
\end{table}

\begin{table}[H]
\centering
\caption{Approximate reproduction compute budget on a single H100 80 GB GPU. Wall-clock ranges depend on model checkpoint, sequence length, and serving environment; VWA additionally uses the official EC2-hosted sites.}
\label{tab:compute}
\small
\begin{tabular}{@{}lcc@{}}
\toprule
\textbf{Experiment} & \textbf{Runs/seeds} & \textbf{Approx. wall-clock} \\
\midrule
WikiText-2 perplexity sweeps & 1 seed, 4 lengths $\times$ 4 models & 2--4 GPU-hours \\
Matched-memory baselines & 1 seed, GPT-2 2048 tokens & 1--2 GPU-hours \\
Mechanistic KL ablation & 1{,}200 sampled steps/model & 2--3 GPU-hours \\
NIAH grid & 3 seeds & 3--6 GPU-hours \\
VisualWebArena & 75 tasks $\times$ 3 seeds & 8--12 GPU-hours plus EC2 environment time \\
Latency/throughput profiling & 3 seeds & 1--2 GPU-hours \\
\bottomrule
\end{tabular}
\end{table}

\clearpage
\section*{NeurIPS Paper Checklist}

\begin{enumerate}

\item {\bf Claims}
    \item[] Question: Do the main claims made in the abstract and introduction accurately reflect the paper's contributions and scope?
    \item[] Answer: \answerYes{}.
    \item[] Justification: The abstract and Introduction state the method, scope, baselines, and empirical claims, and the Results section reports the corresponding measurements.
    \item[] Guidelines:
    \begin{itemize}
        \item The answer \answerNA{} means that the abstract and introduction do not include the claims made in the paper.
        \item The abstract and/or introduction should clearly state the claims made, including the contributions made in the paper and important assumptions and limitations. A \answerNo{} or \answerNA{} answer to this question will not be perceived well by the reviewers. 
        \item The claims made should match theoretical and experimental results, and reflect how much the results can be expected to generalize to other settings. 
        \item It is fine to include aspirational goals as motivation as long as it is clear that these goals are not attained by the paper. 
    \end{itemize}

\item {\bf Limitations}
    \item[] Question: Does the paper discuss the limitations of the work performed by the authors?
    \item[] Answer: \answerYes{}.
    \item[] Justification: Section~\ref{sec:limitations} discusses short-context cases, high-temperature sampling, compute bottlenecks, and compaction overhead.
    \item[] Guidelines:
    \begin{itemize}
        \item The answer \answerNA{} means that the paper has no limitation while the answer \answerNo{} means that the paper has limitations, but those are not discussed in the paper. 
        \item The authors are encouraged to create a separate ``Limitations'' section in their paper.
        \item The paper should point out any strong assumptions and how robust the results are to violations of these assumptions (e.g., independence assumptions, noiseless settings, model well-specification, asymptotic approximations only holding locally). The authors should reflect on how these assumptions might be violated in practice and what the implications would be.
        \item The authors should reflect on the scope of the claims made, e.g., if the approach was only tested on a few datasets or with a few runs. In general, empirical results often depend on implicit assumptions, which should be articulated.
        \item The authors should reflect on the factors that influence the performance of the approach. For example, a facial recognition algorithm may perform poorly when image resolution is low or images are taken in low lighting. Or a speech-to-text system might not be used reliably to provide closed captions for online lectures because it fails to handle technical jargon.
        \item The authors should discuss the computational efficiency of the proposed algorithms and how they scale with dataset size.
        \item If applicable, the authors should discuss possible limitations of their approach to address problems of privacy and fairness.
        \item While the authors might fear that complete honesty about limitations might be used by reviewers as grounds for rejection, a worse outcome might be that reviewers discover limitations that aren't acknowledged in the paper. The authors should use their best judgment and recognize that individual actions in favor of transparency play an important role in developing norms that preserve the integrity of the community. Reviewers will be specifically instructed to not penalize honesty concerning limitations.
    \end{itemize}

\item {\bf Theory assumptions and proofs}
    \item[] Question: For each theoretical result, does the paper provide the full set of assumptions and a complete (and correct) proof?
    \item[] Answer: \answerNA{}.
    \item[] Justification: The paper is an empirical systems paper and does not present theoretical theorems or proofs.
    \item[] Guidelines:
    \begin{itemize}
        \item The answer \answerNA{} means that the paper does not include theoretical results. 
        \item All the theorems, formulas, and proofs in the paper should be numbered and cross-referenced.
        \item All assumptions should be clearly stated or referenced in the statement of any theorems.
        \item The proofs can either appear in the main paper or the supplemental material, but if they appear in the supplemental material, the authors are encouraged to provide a short proof sketch to provide intuition. 
        \item Inversely, any informal proof provided in the core of the paper should be complemented by formal proofs provided in appendix or supplemental material.
        \item Theorems and Lemmas that the proof relies upon should be properly referenced. 
    \end{itemize}

    \item {\bf Experimental result reproducibility}
    \item[] Question: Does the paper fully disclose all the information needed to reproduce the main experimental results of the paper to the extent that it affects the main claims and/or conclusions of the paper (regardless of whether the code and data are provided or not)?
    \item[] Answer: \answerYes{}.
    \item[] Justification: Sections~\ref{sec:method} and~\ref{sec:setup}, plus Appendix~\ref{app:repro}, specify the algorithm, baselines, datasets, hyperparameters, seeds, hardware, and software versions.
    \item[] Guidelines:
    \begin{itemize}
        \item The answer \answerNA{} means that the paper does not include experiments.
        \item If the paper includes experiments, a \answerNo{} answer to this question will not be perceived well by the reviewers: Making the paper reproducible is important, regardless of whether the code and data are provided or not.
        \item If the contribution is a dataset and\slash or model, the authors should describe the steps taken to make their results reproducible or verifiable. 
        \item Depending on the contribution, reproducibility can be accomplished in various ways. For example, if the contribution is a novel architecture, describing the architecture fully might suffice, or if the contribution is a specific model and empirical evaluation, it may be necessary to either make it possible for others to replicate the model with the same dataset, or provide access to the model. In general. releasing code and data is often one good way to accomplish this, but reproducibility can also be provided via detailed instructions for how to replicate the results, access to a hosted model (e.g., in the case of a large language model), releasing of a model checkpoint, or other means that are appropriate to the research performed.
        \item While NeurIPS does not require releasing code, the conference does require all submissions to provide some reasonable avenue for reproducibility, which may depend on the nature of the contribution. For example
        \begin{enumerate}
            \item If the contribution is primarily a new algorithm, the paper should make it clear how to reproduce that algorithm.
            \item If the contribution is primarily a new model architecture, the paper should describe the architecture clearly and fully.
            \item If the contribution is a new model (e.g., a large language model), then there should either be a way to access this model for reproducing the results or a way to reproduce the model (e.g., with an open-source dataset or instructions for how to construct the dataset).
            \item We recognize that reproducibility may be tricky in some cases, in which case authors are welcome to describe the particular way they provide for reproducibility. In the case of closed-source models, it may be that access to the model is limited in some way (e.g., to registered users), but it should be possible for other researchers to have some path to reproducing or verifying the results.
        \end{enumerate}
    \end{itemize}

\item {\bf Open access to data and code}
    \item[] Question: Does the paper provide open access to the data and code, with sufficient instructions to faithfully reproduce the main experimental results, as described in supplemental material?
    \item[] Answer: \answerNo{}.
    \item[] Justification: The submission does not include an anonymized runnable code or data release. The first-page footnote states that code will be released upon acceptance.
    \item[] Guidelines:
    \begin{itemize}
        \item The answer \answerNA{} means that paper does not include experiments requiring code.
        \item Please see the NeurIPS code and data submission guidelines (\url{https://neurips.cc/public/guides/CodeSubmissionPolicy}) for more details.
        \item While we encourage the release of code and data, we understand that this might not be possible, so \answerNo{} is an acceptable answer. Papers cannot be rejected simply for not including code, unless this is central to the contribution (e.g., for a new open-source benchmark).
        \item The instructions should contain the exact command and environment needed to run to reproduce the results. See the NeurIPS code and data submission guidelines (\url{https://neurips.cc/public/guides/CodeSubmissionPolicy}) for more details.
        \item The authors should provide instructions on data access and preparation, including how to access the raw data, preprocessed data, intermediate data, and generated data, etc.
        \item The authors should provide scripts to reproduce all experimental results for the new proposed method and baselines. If only a subset of experiments are reproducible, they should state which ones are omitted from the script and why.
        \item At submission time, to preserve anonymity, the authors should release anonymized versions (if applicable).
        \item Providing as much information as possible in supplemental material (appended to the paper) is recommended, but including URLs to data and code is permitted.
    \end{itemize}

\item {\bf Experimental setting/details}
    \item[] Question: Does the paper specify all the training and test details (e.g., data splits, hyperparameters, how they were chosen, type of optimizer) necessary to understand the results?
    \item[] Answer: \answerYes{}.
    \item[] Justification: Section~\ref{sec:setup} and Appendix~\ref{app:repro} specify models, benchmarks, baselines, decoding settings, hyperparameters, seeds, and hardware. No training optimizer is used because the method is training-free.
    \item[] Guidelines:
    \begin{itemize}
        \item The answer \answerNA{} means that the paper does not include experiments.
        \item The experimental setting should be presented in the core of the paper to a level of detail that is necessary to appreciate the results and make sense of them.
        \item The full details can be provided either with the code, in appendix, or as supplemental material.
    \end{itemize}

\item {\bf Experiment statistical significance}
    \item[] Question: Does the paper report error bars suitably and correctly defined or other appropriate information about the statistical significance of the experiments?
    \item[] Answer: \answerYes{}.
    \item[] Justification: Section~\ref{sec:setup} states that NIAH, VWA, and throughput are reported as mean $\pm$ standard deviation over three seeds; the mechanistic validation reports a correlation and $p$-value.
    \item[] Guidelines:
    \begin{itemize}
        \item The answer \answerNA{} means that the paper does not include experiments.
        \item The authors should answer \answerYes{} if the results are accompanied by error bars, confidence intervals, or statistical significance tests, at least for the experiments that support the main claims of the paper.
        \item The factors of variability that the error bars are capturing should be clearly stated (for example, train/test split, initialization, random drawing of some parameter, or overall run with given experimental conditions).
        \item The method for calculating the error bars should be explained (closed form formula, call to a library function, bootstrap, etc.)
        \item The assumptions made should be given (e.g., Normally distributed errors).
        \item It should be clear whether the error bar is the standard deviation or the standard error of the mean.
        \item It is OK to report 1-sigma error bars, but one should state it. The authors should preferably report a 2-sigma error bar than state that they have a 96\% CI, if the hypothesis of Normality of errors is not verified.
        \item For asymmetric distributions, the authors should be careful not to show in tables or figures symmetric error bars that would yield results that are out of range (e.g., negative error rates).
        \item If error bars are reported in tables or plots, the authors should explain in the text how they were calculated and reference the corresponding figures or tables in the text.
    \end{itemize}

\item {\bf Experiments compute resources}
    \item[] Question: For each experiment, does the paper provide sufficient information on the computer resources (type of compute workers, memory, time of execution) needed to reproduce the experiments?
    \item[] Answer: \answerYes{}.
    \item[] Justification: Appendix~\ref{app:repro} reports the GPU type, memory, CUDA/PyTorch/Transformers versions, the VWA EC2 environment, and approximate wall-clock budgets for each experiment in Table~\ref{tab:compute}.
    \item[] Guidelines:
    \begin{itemize}
        \item The answer \answerNA{} means that the paper does not include experiments.
        \item The paper should indicate the type of compute workers CPU or GPU, internal cluster, or cloud provider, including relevant memory and storage.
        \item The paper should provide the amount of compute required for each of the individual experimental runs as well as estimate the total compute. 
        \item The paper should disclose whether the full research project required more compute than the experiments reported in the paper (e.g., preliminary or failed experiments that didn't make it into the paper). 
    \end{itemize}
    
\item {\bf Code of ethics}
    \item[] Question: Does the research conducted in the paper conform, in every respect, with the NeurIPS Code of Ethics \url{https://neurips.cc/public/EthicsGuidelines}?
    \item[] Answer: \answerYes{}.
    \item[] Justification: The work evaluates inference-efficiency methods on public benchmarks and does not involve human-subject experiments, private data, or new high-risk model releases.
    \item[] Guidelines:
    \begin{itemize}
        \item The answer \answerNA{} means that the authors have not reviewed the NeurIPS Code of Ethics.
        \item If the authors answer \answerNo, they should explain the special circumstances that require a deviation from the Code of Ethics.
        \item The authors should make sure to preserve anonymity (e.g., if there is a special consideration due to laws or regulations in their jurisdiction).
    \end{itemize}

\item {\bf Broader impacts}
    \item[] Question: Does the paper discuss both potential positive societal impacts and negative societal impacts of the work performed?
    \item[] Answer: \answerYes{}.
    \item[] Justification: Section~\ref{sec:limitations} notes both the efficiency benefits of lower inference cost and the risk that cheaper long-context inference can lower barriers to misuse.
    \item[] Guidelines:
    \begin{itemize}
        \item The answer \answerNA{} means that there is no societal impact of the work performed.
        \item If the authors answer \answerNA{} or \answerNo, they should explain why their work has no societal impact or why the paper does not address societal impact.
        \item Examples of negative societal impacts include potential malicious or unintended uses (e.g., disinformation, generating fake profiles, surveillance), fairness considerations (e.g., deployment of technologies that could make decisions that unfairly impact specific groups), privacy considerations, and security considerations.
        \item The conference expects that many papers will be foundational research and not tied to particular applications, let alone deployments. However, if there is a direct path to any negative applications, the authors should point it out. For example, it is legitimate to point out that an improvement in the quality of generative models could be used to generate Deepfakes for disinformation. On the other hand, it is not needed to point out that a generic algorithm for optimizing neural networks could enable people to train models that generate Deepfakes faster.
        \item The authors should consider possible harms that could arise when the technology is being used as intended and functioning correctly, harms that could arise when the technology is being used as intended but gives incorrect results, and harms following from (intentional or unintentional) misuse of the technology.
        \item If there are negative societal impacts, the authors could also discuss possible mitigation strategies (e.g., gated release of models, providing defenses in addition to attacks, mechanisms for monitoring misuse, mechanisms to monitor how a system learns from feedback over time, improving the efficiency and accessibility of ML).
    \end{itemize}
    
\item {\bf Safeguards}
    \item[] Question: Does the paper describe safeguards that have been put in place for responsible release of data or models that have a high risk for misuse (e.g., pre-trained language models, image generators, or scraped datasets)?
    \item[] Answer: \answerNA{}.
    \item[] Justification: The paper does not release a new dataset, model checkpoint, or high-risk generative model; it proposes an inference-time cache-management method.
    \item[] Guidelines:
    \begin{itemize}
        \item The answer \answerNA{} means that the paper poses no such risks.
        \item Released models that have a high risk for misuse or dual-use should be released with necessary safeguards to allow for controlled use of the model, for example by requiring that users adhere to usage guidelines or restrictions to access the model or implementing safety filters. 
        \item Datasets that have been scraped from the Internet could pose safety risks. The authors should describe how they avoided releasing unsafe images.
        \item We recognize that providing effective safeguards is challenging, and many papers do not require this, but we encourage authors to take this into account and make a best faith effort.
    \end{itemize}

\item {\bf Licenses for existing assets}
    \item[] Question: Are the creators or original owners of assets (e.g., code, data, models), used in the paper, properly credited and are the license and terms of use explicitly mentioned and properly respected?
    \item[] Answer: \answerYes{}.
    \item[] Justification: The paper cites all evaluated models, datasets, benchmarks, and baseline methods; Appendix~\ref{app:repro} enumerates their license or terms status in Table~\ref{tab:assets}. We do not redistribute model weights, datasets, benchmark environments, or third-party baseline code.
    \item[] Guidelines:
    \begin{itemize}
        \item The answer \answerNA{} means that the paper does not use existing assets.
        \item The authors should cite the original paper that produced the code package or dataset.
        \item The authors should state which version of the asset is used and, if possible, include a URL.
        \item The name of the license (e.g., CC-BY 4.0) should be included for each asset.
        \item For scraped data from a particular source (e.g., website), the copyright and terms of service of that source should be provided.
        \item If assets are released, the license, copyright information, and terms of use in the package should be provided. For popular datasets, \url{paperswithcode.com/datasets} has curated licenses for some datasets. Their licensing guide can help determine the license of a dataset.
        \item For existing datasets that are re-packaged, both the original license and the license of the derived asset (if it has changed) should be provided.
        \item If this information is not available online, the authors are encouraged to reach out to the asset's creators.
    \end{itemize}

\item {\bf New assets}
    \item[] Question: Are new assets introduced in the paper well documented and is the documentation provided alongside the assets?
    \item[] Answer: \answerNA{}.
    \item[] Justification: The paper does not introduce a new dataset or model asset. An anonymized code package, if submitted, should include separate documentation.
    \item[] Guidelines:
    \begin{itemize}
        \item The answer \answerNA{} means that the paper does not release new assets.
        \item Researchers should communicate the details of the dataset\slash code\slash model as part of their submissions via structured templates. This includes details about training, license, limitations, etc. 
        \item The paper should discuss whether and how consent was obtained from people whose asset is used.
        \item At submission time, remember to anonymize your assets (if applicable). You can either create an anonymized URL or include an anonymized zip file.
    \end{itemize}

\item {\bf Crowdsourcing and research with human subjects}
    \item[] Question: For crowdsourcing experiments and research with human subjects, does the paper include the full text of instructions given to participants and screenshots, if applicable, as well as details about compensation (if any)? 
    \item[] Answer: \answerNA{}.
    \item[] Justification: The work does not involve crowdsourcing or human-subject experiments.
    \item[] Guidelines:
    \begin{itemize}
        \item The answer \answerNA{} means that the paper does not involve crowdsourcing nor research with human subjects.
        \item Including this information in the supplemental material is fine, but if the main contribution of the paper involves human subjects, then as much detail as possible should be included in the main paper. 
        \item According to the NeurIPS Code of Ethics, workers involved in data collection, curation, or other labor should be paid at least the minimum wage in the country of the data collector. 
    \end{itemize}

\item {\bf Institutional review board (IRB) approvals or equivalent for research with human subjects}
    \item[] Question: Does the paper describe potential risks incurred by study participants, whether such risks were disclosed to the subjects, and whether Institutional Review Board (IRB) approvals (or an equivalent approval/review based on the requirements of your country or institution) were obtained?
    \item[] Answer: \answerNA{}.
    \item[] Justification: The work does not involve human subjects, so IRB approval is not applicable.
    \item[] Guidelines:
    \begin{itemize}
        \item The answer \answerNA{} means that the paper does not involve crowdsourcing nor research with human subjects.
        \item Depending on the country in which research is conducted, IRB approval (or equivalent) may be required for any human subjects research. If you obtained IRB approval, you should clearly state this in the paper. 
        \item We recognize that the procedures for this may vary significantly between institutions and locations, and we expect authors to adhere to the NeurIPS Code of Ethics and the guidelines for their institution. 
        \item For initial submissions, do not include any information that would break anonymity (if applicable), such as the institution conducting the review.
    \end{itemize}

\item {\bf Declaration of LLM usage}
    \item[] Question: Does the paper describe the usage of LLMs if it is an important, original, or non-standard component of the core methods in this research? Note that if the LLM is used only for writing, editing, or formatting purposes and does \emph{not} impact the core methodology, scientific rigor, or originality of the research, declaration is not required.
    \item[] Answer: \answerNA{}.
    \item[] Justification: The research studies LLM inference and evaluates LLM checkpoints, which are documented in Sections~\ref{sec:method} and~\ref{sec:setup}. No LLM was used as a non-standard research tool for method development, data generation, or automated evaluation beyond the evaluated models themselves.
    \item[] Guidelines:
    \begin{itemize}
        \item The answer \answerNA{} means that the core method development in this research does not involve LLMs as any important, original, or non-standard components.
        \item Please refer to our LLM policy in the NeurIPS handbook for what should or should not be described.
    \end{itemize}

\end{enumerate}

\end{document}